\documentclass[review]{elsarticle}

\makeatletter
\def\ps@pprintTitle{%
    \let\@oddhead\@empty
    \let\@evenhead\@empty
    \def\@oddfoot{}%
    \let\@evenfoot\@oddfoot
    }
\makeatother

\usepackage{hyperref}

\journal{Neural Networks}

\usepackage{times}
\usepackage{graphicx} 
\usepackage{epsfig} 
\usepackage{subfigure} 

\usepackage{algorithmic}
\usepackage[vlined,ruled,linesnumbered]{algorithm2e}

\usepackage{hyperref}

\usepackage{helvet}
\usepackage{courier}

\usepackage{makecell}
\usepackage{graphicx}
\usepackage{subfigure}
\usepackage{color}
\usepackage{amsfonts}
\usepackage{amsmath}
\usepackage{amssymb}

\usepackage{multirow}
\usepackage{booktabs}

\def\etal{{\em et al.\/}\, }

\def\bmW{\mbox{{\boldmath $\mW$}}}

\def\mX{{\mathcal X}}

\DeclareMathAlphabet\mathbfcal{OMS}{cmsy}{b}{n}

\def\bmO{{\mathbfcal O}}

\def\bmW{{\mathbfcal W}}
\def\bmX{{\mathbfcal X}}

\def\0{{\bf 0}}
\def\1{{\bf 1}}

\def\bG{{\bf G}}

\def\bM{{\bf M}}

\def\bP{{\bf P}}
\def\bQ{{\bf Q}}

\def\bW{{\bf W}}
\def\bX{{\bf X}}
\def\bY{{\bf Y}}

\def\bp{{\bf p}}

\def\bw{{\bf w}}

\def\mmR{{\mathbb R}}

\def\trsp{{\sf T}}

\def\bX{{\bf X}}

\def\bY{{\bf Y}}
\def\bw{{\bf w}}
\def\bW{{\bf W}}

\def\bp{{\bf p}}
\def\bP{{\bf P}}

\def\ie{\mbox{\textit{i.e.}}}
\def\eg{\mbox{\textit{e.g.}}}

\usepackage{ntheorem}

\newtheorem*{*thm}{Theorem}

\newtheorem*{*lemma}{Lemma}

\usepackage{setspace}

\usepackage{hyperref}       
\usepackage{url}            
\usepackage{booktabs}       
\usepackage{amsfonts}       
\usepackage{nicefrac}       
\usepackage{microtype}      
\usepackage{multirow}
\usepackage{graphicx}
\usepackage{amsmath}
\usepackage{arydshln}
\usepackage{diagbox}

\def\cyf{\textcolor{black}}

\begin{document}

\begin{frontmatter}

\title{Content-Aware Convolutional Neural Networks}

\author[mymainaddress,mythirdaryaddress,myfourtharyaddress]{Yong~Guo}
\ead{guo.yong@mail.scut.edu.cn}
\author[mymainaddress]{Yaofo~Chen}
\ead{sechenyaofo@mail.scut.edu.cn}
\author[mymainaddress,mythirdaryaddress]{Mingkui~Tan\corref{mycorrespondingauthor}}
\ead{mingkuitan@scut.edu.cn}
\cortext[mycorrespondingauthor]{Corresponding author}
\author[mymainaddress]{Kui~Jia}
\ead{kuijia@scut.edu.cn}
\author[mymainaddress]{Jian~Chen\corref{mycorrespondingauthor}}
\ead{ellachen@scut.edu.cn}
\author[mymainaddress]{Jingdong~Wang}
\ead{jingdw@microsoft.com}

\address[mymainaddress]{South China University of Technology, China}
\address[mythirdaryaddress]{Key Laboratory of Big Data and Intelligent Robot, Ministry of Education}
\address[myfourtharyaddress]{Pazhou Laboratory, China}
\address[mysecondaryaddress]{Microsoft Research Asia, China}

\begin{abstract}
{Convolutional Neural Networks (CNNs) have achieved great success due to the powerful feature learning ability of convolution layers. Specifically, the standard convolution traverses the input images/features using a sliding window scheme to extract features. However, not all the windows contribute equally to the prediction results of CNNs. In practice, the convolutional operation on some of the windows (\eg, smooth windows that contain very similar pixels) can be very redundant and may introduce noises into the computation. Such redundancy may not only deteriorate the performance but also incur the unnecessary computational cost. Thus, it is important to reduce the computational redundancy of convolution to improve the performance.}
To this end, we propose a Content-aware Convolution (CAC) that automatically detects the smooth windows and applies a $1 \times 1$ convolutional kernel to replace the original large kernel.
{In this sense, we are able to effectively avoid the redundant computation on similar pixels.}
{By replacing the standard convolution in CNNs with our CAC,}
the resultant models yield significantly better performance and lower computational cost than the baseline models with the standard convolution.
More critically, we are able to dynamically allocate suitable computation resources according to the data smoothness of different images, making it possible for content-aware computation.
Extensive experiments on various computer vision tasks demonstrate the superiority of our method over existing methods.
\end{abstract}

\begin{keyword}
Convolution, Neural Networks, Redundancy Reduction.
\end{keyword}

\end{frontmatter}

\section{Introduction}
Recently, convolutional neural networks (CNNs) have achieved remarkable performance in many computer vision tasks, including image classification~\cite{he2016deep,guo2020multi}, face recognition~\cite{schroff2015facenet,sun2015deeply,ozawa2005incremental}, semantic segmentation~\cite{DBLP:journals/pami/ShelhamerLD17,liu2020dynamic,ibtehaz2020multiresunet}, and object detection~\cite{DBLP:journals/pami/RenHG017,wang2018salient}.
Moreover, deep CNNs have also become the workhorse of many other tasks and real-world applications beyond computer vision, such as speech recognition~\cite{skowronski2007automatic,schrauwen2008compact} and natural language processing~\cite{gross2014modeling,duch2008neurolinguistic}.

\begin{figure}[t]
	\centering
	\includegraphics[width=1\textwidth]{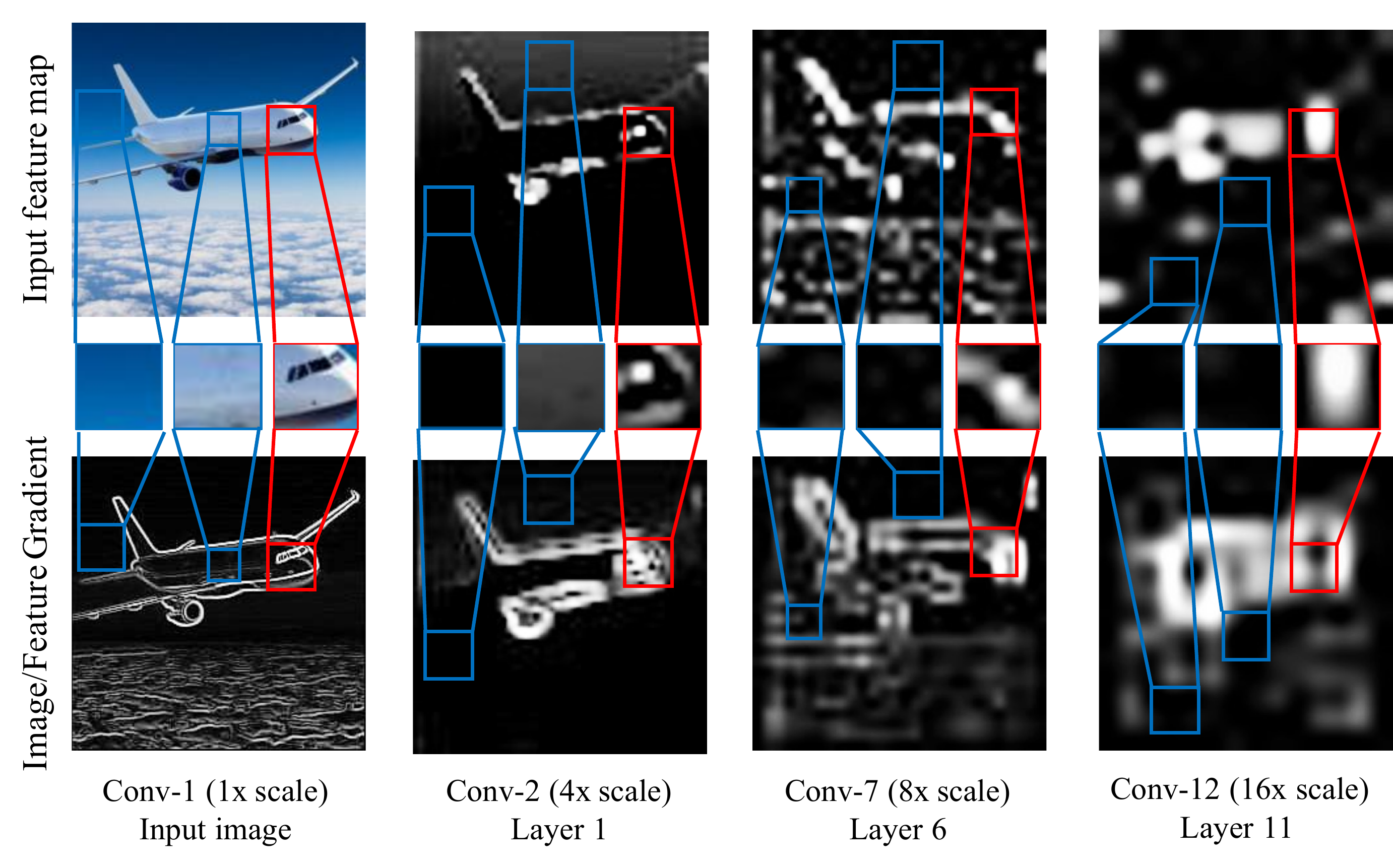}
	\caption{Demonstration of computational redundancy in the input feature maps of different layers of ResNet18 ({pretrained on ImageNet}). The top row and bottom row show the input feature maps and the corresponding gradient {in} different layers, respectively. Red boxes denote the sharp windows that contain the main content of the image. Blue boxes denote the smooth windows that contain limited information about the image and require redundant computation.}
	\label{fig:motivation}
\end{figure}

{One of the key factors behind the success of CNNs lies in the powerful feature learning ability of convolution layers. Typically, the standard convolution transforms the input images/features into a set of windows and exploits a sliding window manner to extract features over them~\cite{burrus1985and}. However, not all the windows contribute equally to the prediction results of CNNs. As shown in Figure~\ref{fig:motivation}, the input images/features often contain a lot of smooth windows that consist of very similar pixels. These windows may contain very limited information about the data~\cite{bar2008efficient,fergus2003object} since a similar pattern may also appear in the surrounding areas. 
As a result, the computation on smooth windows may be very redundant. More critically, performing convolution on these windows may also introduce noises into the computation and thus deteriorate the performance (See results in Tables~\ref{tab:cifar} and~\ref{tab:imagenet}). 
Thus, it is important and necessary to reduce the computational redundancy of convolution to improve the performance.}

Regarding this issue, existing methods improve the convolutional operation by reducing the redundant communications among different channels of feature maps~\cite{krizhevsky2012imagenet,sifre2014rigid,chollet2017xception} or reducing the spatial size of some of the redundant channels~\cite{chen2019drop}.
Specifically, group convolution~\cite{krizhevsky2012imagenet} and depthwise separable convolution~\cite{sifre2014rigid,chollet2017xception} divide the channels of feature maps into multiple groups and perform convolution independently over each group.
Recently, Chen \etal propose the Octave Convolution (OctConv)~\cite{chen2019drop} method, which downscales the feature maps in some redundant channels into smaller sizes to reduce the computational cost. 
However, these methods only {focus on the redundancy inside the channels of feature maps}
but ignore the spatial redundancy of the pixels in each window.
Moreover, existing methods perform the same computation on the samples with different spatial redundancy, which makes the prediction not optimal and also very inefficient.

In this paper, we seek to reduce the computational redundancy on smooth windows to improve the performance of convolution.
To this end, we propose a Content-aware Convolution (CAC) that 
{uses a $1 \times 1$ convolution to replace the computation of original $k \times k$ convolution on smooth windows. In this sense, we are able to effectively avoid the redundant computation to improve the feature learning ability of convolution.}
{To obtain the weights for the $1 \times 1$ convolutional kernels, we spatially aggregating the $k \times k$ kernel by summing up all the kernel parameters (See detailed analysis in Section~\ref{sec:computation_cac}).}
{In order to automatically detect the smooth windows, we propose an effective training method that seeks for a trade-off between model performance and computational cost.}
{More critically}, CAC dynamically allocates computation resources for different samples based on the smoothness of their contents.
Therefore, we are able to perform content-aware computation.
In practice, our CAC models yield significantly better performance and lower computational cost than the models with the standard convolution.
Extensive experiments on different computer vision tasks demonstrate the superiority of our method over existing methods.

In this paper, we make the following contributions.
\begin{itemize}
	\item We propose a Content-aware Convolution (CAC) method that replaces the original $k \times k$ kernel with a $1 \times 1$ kernel on the smooth windows to improve the performance of convolution. 
	With CAC, we are able to effectively reduce the computational redundancy of convolution and significantly improve the performance.
	\item We propose an effective training method to automatically detect the smooth windows for each layer.
	To achieve this goal, we solve a multi-objective optimization problem to find a trade-off between model performance and computational cost.  
    \item Equipped with CAC, the resultant models {achieve content-aware computation by} dynamically allocating computation resources to different samples according to the data smoothness.
	Extensive experiments on different computer vision tasks {demonstrate the effectiveness of the proposed method.}
\end{itemize}

\section{Related Work}\label{sec:related_Studies}

Recently, many efforts have been made to reduce the redundancy of deep networks, including channel pruning, network quantization,
and energy-efficient model design.

\subsection{Channel Pruning} 
Channel pruning is one of the predominant approaches for deep network compression.
Li \etal \cite{li2016pruning} measure the importance of different channels by computing the sum of absolute {values} of weights to conduct channel selection.
Hu \etal \cite{hu2016network} use the average percentage of zeros (APoZ) to select important channels. 
Several training based methods~\cite{alvarez2016learning,liu2017learning} have been proposed to automatically identify the redundant channels by introducing a sparsity regularizer in the training objective.
The reconstruction methods~\cite{He2017channel,luo2017thinet} seek to solve the channel pruning problem by minimizing the reconstruction error between the feature maps of the pretrained model and the compressed model.
Recently, Zhuang \etal \cite{zhuang2018discrimination} propose a discrimination-aware channel pruning (DCP) method to choose the channels that contribute to the discriminative power and obtain state-of-the-art results.
{Based on DCP, Zhuang \etal \cite{liu2020discrimination} further propose a discrimination-aware kernel pruning (DKP) method by removing the redundant kernels according to the discrimination power.}
However, these methods only focus on the redundancy in model parameters but ignore the redundancy incurred by the input data.

\subsection{Network Quantization}
Network quantization aims to convert the pretrained full-precision convolution networks into the low-precision versions to reduce the computational cost.
Recently, Song \etal \cite{han2015deep} propose a three-stage deep compression pipeline, including pruning, trained quantization, and Huffman coding.
DoReFa-Net~\cite{zhou2016dorefa} seeks to quantize the full precision weights, activations, and gradients to the low bit ones for deep networks. 
In ternary weight networks (TWNs)~\cite{li2016ternary,zhu2016trained}, the parameters are constrained to $+1$, $0$, and $-1$, and {the model achieves} higher accuracy than binary neural networks.
Similar to channel pruning, network quantization also reduces the redundancy in model parameters and may yield limited performance.

\subsection{Energy-efficient Model Design} 

Many energy-efficient modules have been proposed to reduce the computational cost of deep networks. 
Specifically, sparse convolution~\cite{Liu_2015_CVPR,graham2017submanifold} zeros out a large number of parameters to reduce the model size.
Group convolution~\cite{krizhevsky2012imagenet} and depthwise separable convolution~\cite{sifre2014rigid} divide the input channels into groups to reduce the redundant communications among different groups. NAT~\cite{guo2019nat,guo2021towards} replaces redundant operations with identity mapping or directly remove them to obtain efficient models.

Related to our method, Li \etal \cite{li2017not} propose a Region Convolution (RC) that reduces the computational redundancy for semantic segmentation models.
Specifically, given an input feature map, RC performs convolutions on the regions of hard pixels and discards the easy pixels according to the confidence of the predicted mask. However, it has some underlying limitations.
First, RC relies on the predicted confidence of all pixels to construct the mask and cannot be applied to tasks without dense prediction, \eg, image classification. 
Second, RC completely discards the easy regions and may influence the features learned in the deeper layers.
Unlike RC, our CAC preserves the information in all regions/pixels and can be applied to most computer vision tasks, \eg, image classification, semantic segmentation, and object detection.

Very recently, Chen \etal propose the octave convolution (OctConv)~\cite{chen2019drop} method, which reduces the spatial resolution of some low-frequency feature maps to reduce the computational complexity.
However, it has two major limitations.
First, the low-frequency feature maps are predefined before training rather than detected according to the input data.
Second, OctConv only considers the redundancy in the channels of the feature maps but ignores the redundancy in the data content, \eg, pixels.
Compared to OctConv, our CAC automatically detects the sharp/smooth windows from the input data to achieve content-aware computation.
Moreover, CAC considers the pixel level redundancy caused by the input data.

\section{Notations and Problem Definition}\label{sec:notations}
In this paper, we assume that the convolution has the stride of 1 and is performed with padding to guarantee that the output feature maps have the same size as the input feature maps. 
In this paper, we consider the squared input images and feature maps.
For simplicity, we consider one single-channel input feature map $\bX \in \mmR^{n \times n}$ and one convolutional kernel $\bW \in \mmR^{k \times k}$, where $n$ and $k$ denote the sizes of the feature maps and kernels, respectively. Thus, the standard convolution can be written as 
\begin{equation}\label{eq:convolution}
\bY = \bX \otimes \bW,
\end{equation}
where $\otimes$ denotes the convolutional operator.

\begin{figure*}[t]
	\centering
	\includegraphics[width=0.95\textwidth]{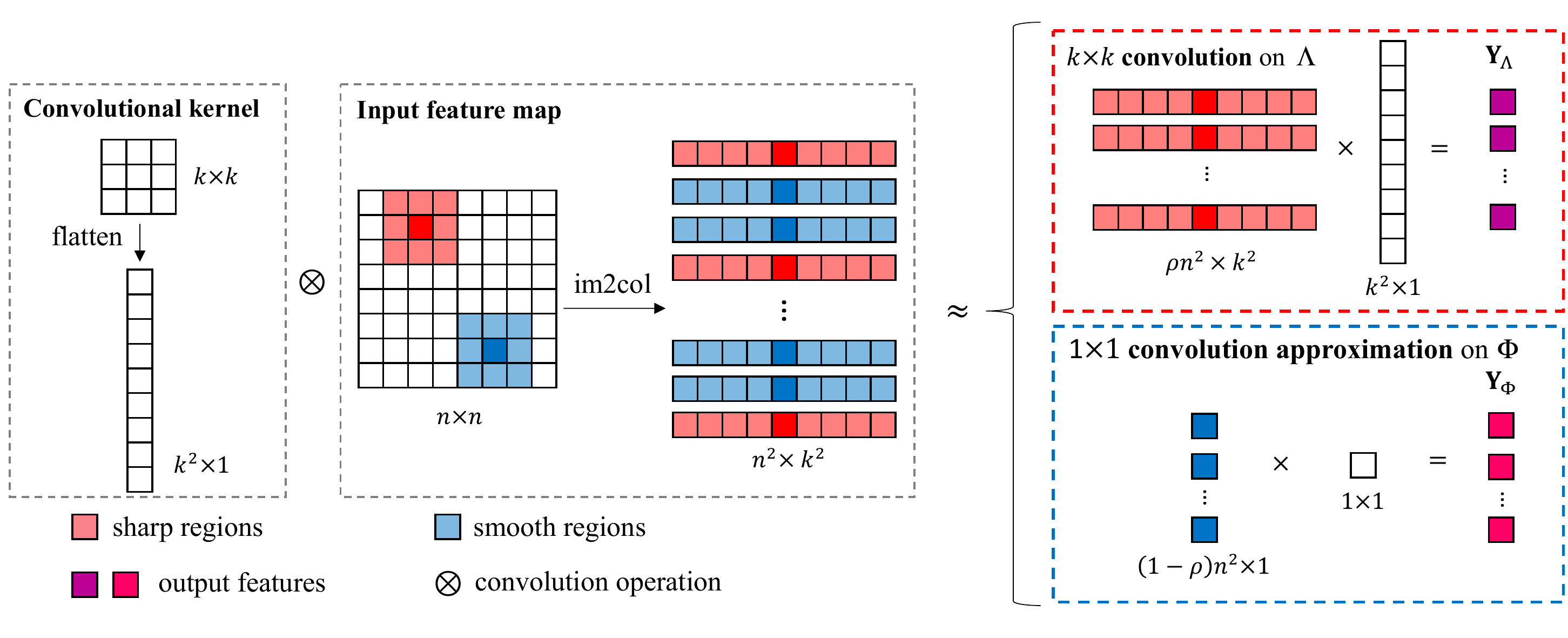}
	\caption{The computation method of Content-aware Convolution.
	{We first divide the input feature maps into two parts, namely the sharp windows (red boxes) and smooth windows (blue boxes). Then, we perform $k \times k$ convolution on sharp windows and $1 \times 1$ convolution on smooth windows. 
	}}
	\label{fig:architecture}
\end{figure*}

In practice, the computation of convolution is often converted to the matrix-matrix or matrix-vector multiplication~\cite{vasudevan2017parallel}.
Given a convolution {with a $k \times k$ kernel} and an input feature map $\bX \in \mmR^{n \times n}$, there are $n^2$ windows that are convolved by the kernel~\cite{ludwig2013image}. 
In this sense, we can represent $\bX$ by a set of windows
\begin{equation}\label{eq:set_I}
\Psi := \{ \bQ_i \in \mmR^{k \times k} {~|~i = 1,...,n^2} \},
\end{equation}
where $\bQ_i$ denotes the $i$-th window in $\Psi$.
For any window $\bQ_i$, we can reshape it into a vector $\bp_i = \mathrm{vec}(\bQ_i)\in \mmR^{k^2}$.
For convenience, we define $\bP := [\bp_1,..., \bp_{n^2}] \in \mmR^{k^2\times n^2}$ and $\bw :=  \mathrm{vec}({\bW})\in \mmR^{k^2}$.  
Note that the transformation from $\bX$ to $\bP$ is often called im2col (See Fig.~\ref{fig:architecture}).
Thus, Eqn.~(\ref{eq:convolution}) can be written as a matrix-vector multiplication:
\begin{equation}\label{eq:matmulti}
\begin{aligned}
\bY = \mathrm{vec2mat}(\bP^{\trsp} \bw), ~~~
\end{aligned}
\end{equation}
where the function $\mathrm{vec2mat}(\cdot)$ denotes the operation to reshape a vector to a matrix.

Clearly, the complexity of the matrix-vector multiplication is $O(n^2k^2)$, which can be very expensive when $n$ and/or $k$ are very large.
However, some of the windows are very smooth and only contain limited information.
{More critically, performing convolution on smooth windows may also introduce noises into the computation and thus hamper the performance.}
Thus, it is necessary and important to reduce the computational redundancy on smooth windows to improve the performance of convolution.

\section{Proposed Method}\label{sec:method}
In this paper, we propose a Content-aware Convolution (CAC) {that replaces the original kernel with a $1 \times 1$ kernel to perform convolution on smooth windows. Moreover, we present an effective method to automatically detect smooth windows.} 
We show the overall scheme in Fig.~\ref{fig:architecture} and the detailed computation method of CAC in Algorithm~\ref{alg:cacc}.

\subsection{Motivation}~\label{sec:motivation}
The standard convolution convolves the input images/features by transforming them into a set of windows and adopts a sliding window scheme to extract features~\cite{ludwig2013image}. 
However, not all the windows contribute equally to the prediction results of deep networks.
As shown in Fig.~\ref{fig:motivation}, there are a large number of smooth windows in the feature maps of each layer.
These windows often contain very similar pixels and come with very limited information about the data.
{As a result, performing convolution on smooth windows can be redundant. More critically, the computational redundancy on these smooth windows may also introduce noises into the computation and thus deteriorate the performance.}

Instead of convolving all the windows using the same kernel, {we seek to perform different convolutional computations on the windows according to their smoothness. Specifically, we first recognize the sharp and smooth windows.}
Then, we perform the standard convolution on the sharp windows and perform the convolution with a smaller kernel of $1 \times 1$ on a single pixel of the smooth windows to reduce the computational redundancy. 
Since the computation depends on the content of the input data {in} each layer, we call our method \textbf{Content-aware Convolution (CAC)}.

\begin{algorithm}[t]
	\caption{\small Content-Aware Convolution (CAC).}
    	\begin{algorithmic}[1]\small
    		\REQUIRE  Input feature map $\bX \in \mmR^{n \times n}$;\\
    		~~~~~~~Convolutional kernel $\bW \in \mmR^{k \times k}$;\\
    		~~~~~~~Learnable parameters kernel $\gamma$ and $\beta$;\\
    		~~~~~~~Set of input windows $\Psi := \{ \bQ_i \in \mmR^{k \times k} {~|~i = 1,...,n^2} \}$.\\
    		\STATE Compute the $1 {\small \times} 1$ kernel $w_{\Phi}$ using Eqn.~(\ref{eq:kernel_approximation});\\
    		\STATE Compute the average feature map using Eqn.~(\ref{eq:average});\\
    		\STATE Compute the gradient of feature maps $\bG$ using Eqn.~(\ref{eq:gradient});\\
    		\STATE Compute the score map based on $\bG$: \\ 
    		~~~~~~~~~~~~~~~~~~~$\bM = {\rm Sigmoid}~ (\gamma \bG + \beta)$;\\
    		\STATE Obtain the set of sharp windows: \\
    		~~~~~~~~~~~~~~~~~~~~~$\Lambda = \{ \bQ_{i} ~|~  M_{i} > 0.5 \}$;\\
    		\STATE Obtain the set of smooth windows:\\
    		~~~~~~~~~~~~~~~~~~~~~~~~~~$\Phi = \Psi ~\backslash~ \Lambda$;\\
    		\STATE Perform convolution on $\Lambda$: \\
    		~~~~~~~~~~~~~~~~~~~~~~~$\bY_{\Lambda} = {\rm Conv} (\Lambda; \bW)$;\\
    		\STATE Perform convolution on $\Phi$: \\
    		~~~~~~~~~~~~~~~~~~~~~~~$\bY_{\Phi} = {\rm Conv} (\Phi; w_{\Phi})$;\\
    		\STATE Combine $\bY_{\Lambda}$ and $\bY_{\Phi}$ to obtain the final output: \\
    		~~~~~~~~~~~~~~~~~~~~~$\bY = {\rm Combine} (\bY_{\Lambda}, \bY_{\Phi})$.\\
    	\end{algorithmic}
	\label{alg:cacc}
\end{algorithm}

\subsection{Content-aware Convolution}\label{sec:computation_cac}
Given an input feature map or image, we divide the whole window set $\Psi$ into two disjoint subsets, namely the \textbf{sharp window set} $\Lambda$ and the \textbf{smooth window set} $\Phi$.
We will illustrate how to detect smooth/sharp windows from $\Psi$ in Section~\ref{sec:region}.
{To reduce the computational redundancy on smooth windows,}
we seek to use a $1 \times 1$ kernel to replace the original large kernel. 
Given an input window $\bQ_i \in \mmR^{k \times k}$ and a single-channel kernel $\bW \in \mmR^{k \times k}$, the output of a convolution layer $y_i$ can be computed by
\begin{eqnarray}\label{eq:window-conv}
y_i = \bQ_i \otimes \bW = \bp^{\top}_{i} \bw = \sum_{j=1}^{k^2} p_{j} w_j,
\end{eqnarray}
where $\bp_{i}$ denotes vector presentation of $\bQ_i$ and $p_{j}$ denotes the $j$-th element of $\bp_i$.
If $\bQ_i$ is a smooth window, it {implies} that all the elements of the window should have very similar values, \ie, for $\forall m,n \small{\in} \{1,...,k^2\}, p_{m}  \approx p_{n}$.
Therefore, it follows that
	\begin{equation}\label{eq:conv_approximation}
	\begin{aligned}
	\bp^{\top}_{i} \bw = \sum_{j=1}^{k^2} p_{j} w_j  \approx \bar{p} \cdot \sum_{j=1}^{k^2} w_j,
	\end{aligned}
	\end{equation}
where $\bar{p}$ can be the average value of all $p_j$ $(\bar{p} = \frac{1}{k^2} \sum_{j=1}^{k^2} p_{j})$
or any element of this window. {In this paper}, we choose the center element to compute $\bar{p}$.
Relying on Eqn.~(\ref{eq:conv_approximation}),
we approximate the original $k \times k$ convolution kernel using a $1 \times 1$ kernel:
\begin{equation}\label{eq:kernel_approximation}
w_{\Phi} = \sum_{j=1}^{k^2} w_j.
\end{equation}
{Note that the computation on similar pixels may introduce noises into the computation. Our CAC performs convolution on a single pixel in a window and thus effectively reduces the impact of the noisy information. In this sense, the CAC based models are often more robust than the models built with the standard convolution (See results in Table~\ref{tab:robustness}).}
{Moreover, the computational cost of CAC on smooth windows can be reduced to ${1}/{k^2}$ of the cost with the $k \times k$ kernels.}

Given the sharp windows $\Lambda$ and smooth windows $\Phi$, we obtain the output of CAC by performing a $k \times k$ convolution and a $1 \times 1$ convolution on  $\Lambda$ and $\Phi$, respectively. 
Let $\bW \in \mmR^{k \times k}$ be the parameters of a {single-channel} convolutional kernel, $w_{\Phi}$ be the $1 \times 1$ kernel obtained by Eqn.~(\ref{eq:kernel_approximation}). The computation on $\Lambda$ and $\Phi$ can be formulated by
\begin{equation}\label{eq:seperate_conv}
		\bY_{\Lambda} = {\rm Conv} (\Lambda; \bW), ~~\bY_{\Phi} = {\rm Conv} (\Phi; w_{\Phi}).
\end{equation}
Then, we combine $\bY_{\Lambda}$ and $\bY_{\Phi}$ to obtain the final output according to the relative positions of the windows.
\begin{equation}\label{eq:combined_conv}
    \bY = {\rm Combine} (\bY_{\Lambda}, \bY_{\Phi}).
\end{equation}

\subsection{Sharp and Smooth Window Recognition}\label{sec:region}
Based on the smoothness of windows, we propose an effective method to automatically detect the sharp/smooth windows.
{In this paper, we measure the data smoothness using the gradients of the input images or features.}

For any layer of a deep network, not all the channels are useful and some of them are noisy or irrelevant to the final prediction results~\cite{zhuang2018discrimination}. 
{As a result, the features in these channels can be very noisy and thus may hamper the final prediction results~\cite{wang2015visual}.}
Regarding this issue, we seek to compute the average feature map over different channels to alleviate the influence of noisy features:
\begin{equation}\label{eq:average}
	\overline \bX = \frac{1}{m} \sum_{i=1}^{m} \bX_i,
\end{equation}
where $m$ denotes the number of channels.
Then, we compute the gradient of the averaged feature map to compare the sharpness of different windows.
In this paper, we use the Sobel operator~\cite{sobel} to compute the gradient by performing two 1-d convolutions along the x- and y-axis, respectively:
\begin{equation}\label{eq:gradient_xy}
\nonumber
\bG_x =  
\overline \bX
\otimes
\left[
\begin{gathered}
\small{-}1 ~~0 ~~\small{+}1 \\
\end{gathered}
\right]
\otimes
\left[
\begin{gathered}
1~ \\
2~ \\
1~ \\
\end{gathered} \right],~~~
\bG_y =
\overline \bX
\otimes
\left[
\begin{gathered}
\small1 ~~2 ~~1 \\
\end{gathered}
\right]
\otimes
\left[
\begin{gathered}
\small {-}1~ \\
0~ \\
\small{+}1~ \\
\end{gathered} \right].
\end{equation}
Thus, the total gradient becomes
\begin{equation}\label{eq:gradient}
\bG = \sqrt {\bG_x^2 + \bG_y^2}.
\end{equation}
{Given the feature map with $m$ channels, computing gradients on the averaged feature map only yields $1/m$ cost of the computation on all the channels. Compare to the cost of convolution, the cost of computing gradients can be negligible in practice.}

Based on the computed gradients, we may divide the windows into sharp and smooth windows {according to some threshold}.
However, such a threshold has to be carefully selected for each layer, making it very time consuming and labor-intensive.
To address this issue, we propose a learnable module to automatically discriminate the sharp windows from the smooth windows.
Specifically, we exploit an affine function to {transform the gradients} and then apply the Sigmoid function to compute the probability of a window being sharp. The probability map $\bM$ can be computed by
\begin{equation}\label{eq:affine}
	\bM = {\rm Sigmoid}~ (\gamma \bG + \beta),
\end{equation}
where $\gamma$ and $\beta$ are trainable parameters. 

Here, we consider the windows with the probability larger than 0.5 as sharp windows {and the other windows as smooth windows. Formally, the sets of sharp and smooth windows can be represented by}:
\begin{equation}\label{eq:set_P}
\Lambda = \{ \bQ_{i} ~|~  M_{i} > 0.5 \}, ~~~\Phi = \Psi ~\backslash~ \Lambda,
\end{equation}
where $M_i$ is the score of the window $\bQ_i$ and $\Psi$ denote the set of all the windows.
By changing a hard threshold manner to a learnable scheme, the model is able to adjust $\gamma$ and $\beta$ to find the optimal number of smooth windows to perform $1 \times 1$ convolution.
We will show the training method for $\gamma$ and $\beta$ in Section~\ref{sec:training}.

\subsection{Training Method}\label{sec:training}

Note that CAC seeks to find a number of smooth windows to perform $1 \times 1$ convolution to improve the performance. Although we can reduce the computational cost, performing $1 \times 1$ convolution on too many windows may also hamper the performance (See results in Table~\ref{tab:lambda}).  
To find a good trade-off between model performance and computational cost, we propose to solve a multi-objective optimization problem.

Let $M$ be the CAC-based model to be trained, $M^b$ be the baseline model with the standard convolution, and $c(*)$ be the function to measure the computational cost of deep models, \eg, the number of multiply-adds (MAdds). 
To train CAC-based models, we use the weighted product method\footnote{We use the weighted product method because it is easy to customize for different models. The weighted sum method is also appropriate.} to build the objective:
\begin{equation}\label{eqn:loss}
	{L} = \ell(M) \left( \frac{c(M)}{c(M^b)} \right)^{\lambda},
\end{equation}
where $\ell(M)$ denotes the standard loss function w.r.t. $M$ (\eg, the cross-entropy loss for classification models) and $\lambda \geq 0$ is a constant weight factor.
When $\lambda=0$, the objective is reduced to the standard loss for a specific task. When $\lambda > 0$, 
we seek to find a promising trade-off between model performance and computational cost.
In this paper, to obtain a good balance, we use $\lambda$ to control the importance of computational cost (See discussions on $\lambda$ in Section~\ref{exp:lambda}).

\section{More Discussions}

\cyf{In this section, we first analyze the computational complexity of the proposed CAC convolution in Section~\ref{sec:complexity}.
Then, we discuss the differences between the proposed methods and existing methods in Section~\ref{sec:difference}.}

\subsection{Computational Complexity Analysis}\label{sec:complexity}

To analyze the computational complexity of the proposed CAC, we consider the more general case in which a convolution layer contains multiple channels. Let $\pmb{\mX} \in \mmR^{n \times n \times c_{in}}$ be the input feature maps of a convolution layer and the convolutional kernel be
$\bmW \in \mmR^{k \times k \times c_{in} \times c_{out}}$, where $c_{in}$ and $c_{out}$ are the numbers of input and output channels, respectively. {Then, the convolution in a standard convolution layer can be computed by}
\begin{equation}\label{eq:tensor_convolution}
\bmO = \bmX \otimes \bmW, ~~~\bmO \in \mmR^{n \times n \times c_{out}},
\end{equation}
where $\otimes$ denotes the convolution operation.
The number of multiply-adds (MAdds) required by the standard convolution is given by:
\begin{equation}\label{eq:complexity_conv}
\Omega_{\rm Conv} = c_{in} \cdot c_{out} \cdot k \cdot k \cdot n \cdot n.
\end{equation}

Given a specific proportion of sharp windows (denoted by $\rho$), the computational complexity consists of three parts. \textbf{First},
there are $\rho \cdot n^2$ sharp windows in $\Lambda$ where we perform the standard convolution using the $k \times k$ kernel.
Thus, the complexity of the first part becomes $\rho\Omega_{\rm conv}$. 
\textbf{Second}, we perform $1 \times 1$ convolution on $(1\small{-}\rho)\cdot n^2$ smooth windows in $\Phi$, each of which only requires ${1}/{k^2}$ complexity of the standard $k \times k$ convolution. Therefore, the computational complexity of the second part is $(1 - \rho)\Omega_{\rm conv}/k^2 .$ 
\textbf{Third}, we perform two 1-d convolutions with {a} single output channel to compute the gradients along the x- and y-axis, respectively. Thus, there are a total of four $1 \times 3$ or $3 \times 1$ convolutions to compute the gradient.

Compared to the standard convolution with $c_{in}$ input channels and $c_{out}$ output channels, the complexity of computing gradients in the third part is
\begin{equation}
\underbrace{4 \cdot \frac{3}{k^2} \cdot \frac{1} {c_{in} c_{out}} \Omega_{\rm Conv}}_{\text{computing gradient}} + \underbrace{\frac{1}{k^2 c_{in}c_{out}} \Omega_{\rm conv}}_{\text{linear transformation}}
= \frac{13}{k^2 c_{in}c_{out}} \Omega_{\rm conv}.
\end{equation}
As a result, the total computational complexity of the CAC convolution becomes:
\begin{equation}\label{eq:complexity_fast}
\Omega_{\rm CAC} = \left( \underbrace{\rho}_{\text{$k \times k$ conv}} + \underbrace{\frac{1-\rho}{k^2}}_{\text{$1 \times 1$ conv}} +  \underbrace{\frac{13}{k^2 c_{in} c_{out}}}_{\text{computing score map}} \right) \Omega_{\rm Conv}.
\end{equation}

To accelerate the computation of the convolutions, according to Eqn.~(\ref{eq:complexity_fast}), we have to satisfy the condition such that ${\Omega_{\rm CAC}} / {\Omega_{\rm Conv}}\leq 1$. In this sense, we can obtain the upper bound of the ratio $\rho$:
\begin{equation}\label{eq:acce_condition}
\rho \leq \overline{\rho} = 1 - \frac{13}{(k^2 - 1) \cdot c_{in} \cdot c_{out}}.
\end{equation}

Specifically, for a $3 \times 3$ convolution (\ie, $k=3$), the upper bound is $\overline{\rho} = 1 - \frac{13}{8 \cdot c_{in} \cdot c_{out}}$. 
Taking ResNet~\cite{he2016deep} as an example, the number of channels $c_{out}$ ranges from 16 to 512. In this sense, the ratio only needs to be $\rho < 99.3\%$ when we substitute the smallest value $c_{in}=c_{out}=16$ into Eqn.~(\ref{eq:acce_condition}). Thus, there is considerable potential to accelerate the computation of the standard convolution.

\subsection{{Differences from Existing Methods}}\label{sec:difference}

\cyf{
The proposed CAC method has several essential differences from existing methods.
\textbf{First}, the standard convolution performs convolution using a general kernel on all the windows and ignores the inherent redundancy, which may hamper the performance (See results in Tables~\ref{tab:cifar} and~\ref{tab:imagenet}). In contrast, the proposed CAC performs different convolutions on these windows to reduce the computational redundancy and improve the performance.
\textbf{Second}, existing methods perform the same computation on the samples with different spatial redundancy. Unlike these methods, our CAC adopts a content-aware computation scheme that dynamically allocates suitable computational resources for different samples according to their data smoothness. 
It is worth noting that our CAC is more robust to the samples with adversarial perturbations than existing convolution methods. The main reason is that CAC replaces the large kernel convolution with a $1 \times 1$ convolution on smooth windows, which effectively reduces the influence incurred by the noises/attacks in these windows.
}

\section{Experiments}\label{sec:exp}

In this section, we use CAC to accelerate two popular convolution methods, namely the standard convolution and the octave convolution (OctConv)~\cite{chen2019drop}.
We apply CAC to various architectures and demonstrate the performance on three computer vision tasks, including image classification, semantic segmentation, and object detection.
All implementations are based on PyTorch\footnote{The implementation of the proposed CAC method is available at \href{https://github.com/guoyongcs/CAC}{https://github.com/guoyongcs/CAC}.}.

We organize the experiments as follows.
First, we show the visual interpretation of each CAC layer in Section~\ref{sec:feature_visualization}.
Second, we evaluate our CAC on image classification tasks in Section~\ref{sec:image_classification}.
Third, we apply our CAC method to semantic segmentation models and evaluate the proposed method in Section~\ref{sec:semantic_segmentation}.
Fourth, we conduct experiments to show the effectiveness of our CAC method on object detection tasks in Section~\ref{sec:object_detection}. Finally, we investigate the effect of the hyperparameter $\lambda$ in Section~\ref{exp:lambda}.

\begin{figure*}[t]
	\centering
	\includegraphics[width=1\textwidth]{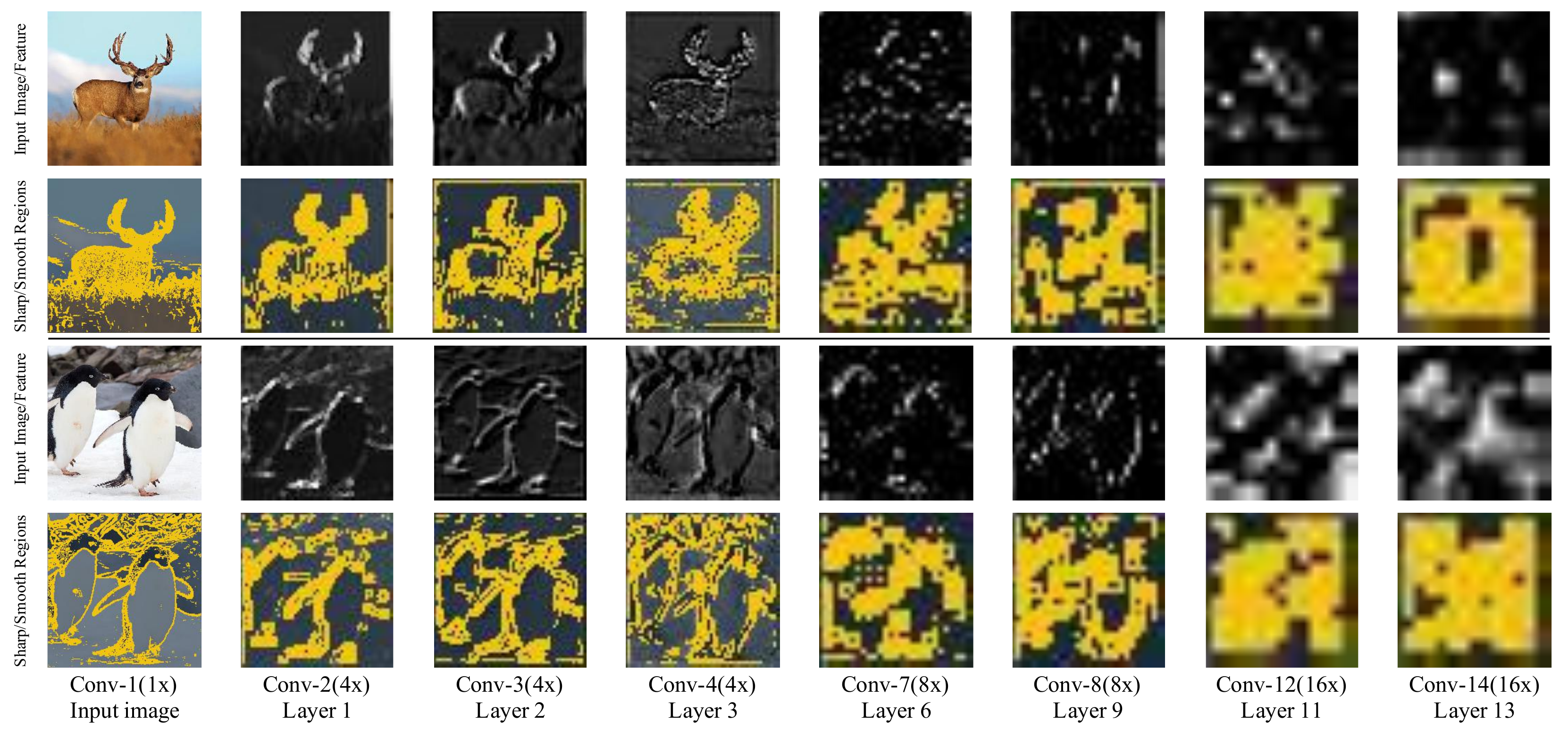}
	\caption{Visualization of the feature maps of different layers in {CAC-ResNet18} on ImageNet. For each image, the top row shows the feature map of different layers and the bottom row shows the corresponding map of sharp windows detected by CAC. In the bottom row, yellow regions denote the sharp windows to perform $3 \times 3$ convolution and the dark regions denote the smooth ones to perform $1 \times 1$ convolution. We scale the feature maps of different layers to the same spatial size for better visualization.}
	\label{fig:feature}
\end{figure*}

\subsection{Visual Interpretation of CAC Convolution}\label{sec:feature_visualization}

To better understand the proposed CAC method, we visualize the feature maps and the corresponding masks of the sharp windows of different CAC layers inside deep networks.
In this experiment, we take the CAC-ResNet18 model as an example and show the results in Fig.~\ref{fig:feature}.

From Fig.~\ref{fig:feature}, the sharp windows (marked in yellow) are often located at the edges and contain the main information about the object.
However, the smooth windows (marked in black) are often very smooth areas that only contain little information.  
As discussed in Section~\ref{sec:method}, when the input windows are very smooth, it is not necessary to use a large kernel to perform convolution. 
Thus, we can reduce the computational cost of convolution on the smooth windows using a $1 \times 1$ kernel to approximate the original output.
In this way, our CAC method can greatly reduce the computational complexity without a loss of information. 
More critically, since the input images or features may have different numbers of smooth windows, the resultant CAC models can dynamically allocate suitable computation power to different input images.
Thus, our CAC models can perform content-aware computation to improve the performance of convolution.

\subsection{Experiments on Image Classification}\label{sec:image_classification}

In this experiment, we consider two popular convolution methods as the baseline methods, namely, the standard convolution and the OctConv~\cite{chen2019drop}.
We apply the proposed CAC method to various image classification models, including ResNet~\cite{he2016deep}, DenseNet~\cite{huang2017densely}, and ShuffleNetV2~\cite{ma2018shufflenet}.

\subsubsection{Datasets and Implementation Details}

{We conduct experiments on two benchmark image classification datasets, including CIFAR-10~\cite{krizhevsky2009learning}, and ImageNet~\cite{deng2009imagenet}.}
CIFAR-10 consists of 50k training samples and 10k testing images with 10 classes. ImageNet contains 123k training samples and 50k testing images for 1,000 classes.

We follow the settings in~\cite{he2016deep} and use SGD with nesterov~\cite{nesterov1983method} for the optimization. The momentum and weight decay are set to 0.9 and 0.0001, respectively.
On CIFAR-10, we train the models for 400 epochs using a mini-batch size of 128. 
The learning rate is initialized to 0.1 and divided by 10 at epochs 160 and 240, respectively. 
On ImageNet, we train the models for 90 epochs with a mini-batch size of 256. 
The learning rate is started at 0.1 and divided by 10 at epochs 30 and 60, respectively.
We train the CAC based models with $\lambda=0.3$. 
We use the number of multiply-adds (MAdds) to measure the computational complexity of deep models. 
Based on sharp window set $\Lambda$ and the set of all the windows $\Psi$, we compute the ratio of sharp windows inside a convolution layer by $ \rho = | \Lambda | / | \Psi | $, where $|\cdot|$ denotes the cardinality of a set.
In general, a lower ratio implies that the more windows would be convolved with a $1 \times 1$ kernel.

\begin{table*}[t!]
  \centering
  \caption{Comparisons of different convolutions in terms of both computational complexity and testing error based on various architectures on CIFAR-10. ``$\backslash$'' denotes the missing results of the models to which OctConv cannot be applied.}
  \resizebox{\textwidth}{!}
  {
    \begin{tabular}{c||cc|cc|cc|cc|cc}
    \hline
    \multirow{2}[0]{*}{Conv Type} & \multicolumn{2}{c|}{ResNet20} & \multicolumn{2}{c|}{ResNet32} & \multicolumn{2}{c|}{ResNet56} & \multicolumn{2}{c|}{DenseNet121} & \multicolumn{2}{c}{ShuffleNetV2} \\
    & \multicolumn{1}{c}{\#MAdds} & \multicolumn{1}{c|}{Error} & \multicolumn{1}{c}{\#MAdds} & \multicolumn{1}{c|}{Error} & \multicolumn{1}{c}{\#MAdds} & \multicolumn{1}{c|}{Error} & \multicolumn{1}{c}{\#MAdds} & \multicolumn{1}{c|}{Error} & \multicolumn{1}{c}{\#MAdds} & \multicolumn{1}{c}{Error} \\
     & (M) & (\%) & (M) & (\%) & (M) & (\%) & (M) & (\%) & (M) & (\%) \\
    \hline
    Standard Conv  & 40.93 & 8.75 & 69.12 & 7.51 & 126.08 & 6.97 & 888.51 & 4.78 & 45.04 &7.25 \\
    CAC  & \textbf{25.79} & \textbf{8.37} & \textbf{46.58} & \textbf{7.25}    & \textbf{87.27} & \textbf{6.51} & \textbf{733.82} & \textbf{4.60} & \textbf{41.33} & \textbf{7.13} \\
    \hline
    OctConv~\cite{chen2019drop} & 26.33 & 8.77 & 44.02 & 8.07 & 61.72 & 7.58 & 403.80 & 5.52 & \multicolumn{2}{c}{\multirow{2}[0]{*}{\diagbox{~~~~~~~~~~}{~~~~~~~~~~}}}   \\
    CAC-OctConv & \textbf{19.49} & \textbf{8.60} & \textbf{34.39} & \textbf{7.63} & \textbf{52.12} & \textbf{6.99} & \textbf{356.63} & \textbf{4.68} & &    \\   
    \hline
    \end{tabular}%
  }
  \label{tab:cifar}%
\end{table*}%

\subsubsection{Comparisons on CIFAR-10}

\begin{figure*}[t!]
	\centering
	\includegraphics[trim = 1mm 0mm 0mm 0mm,
	clip, width=0.9\textwidth]{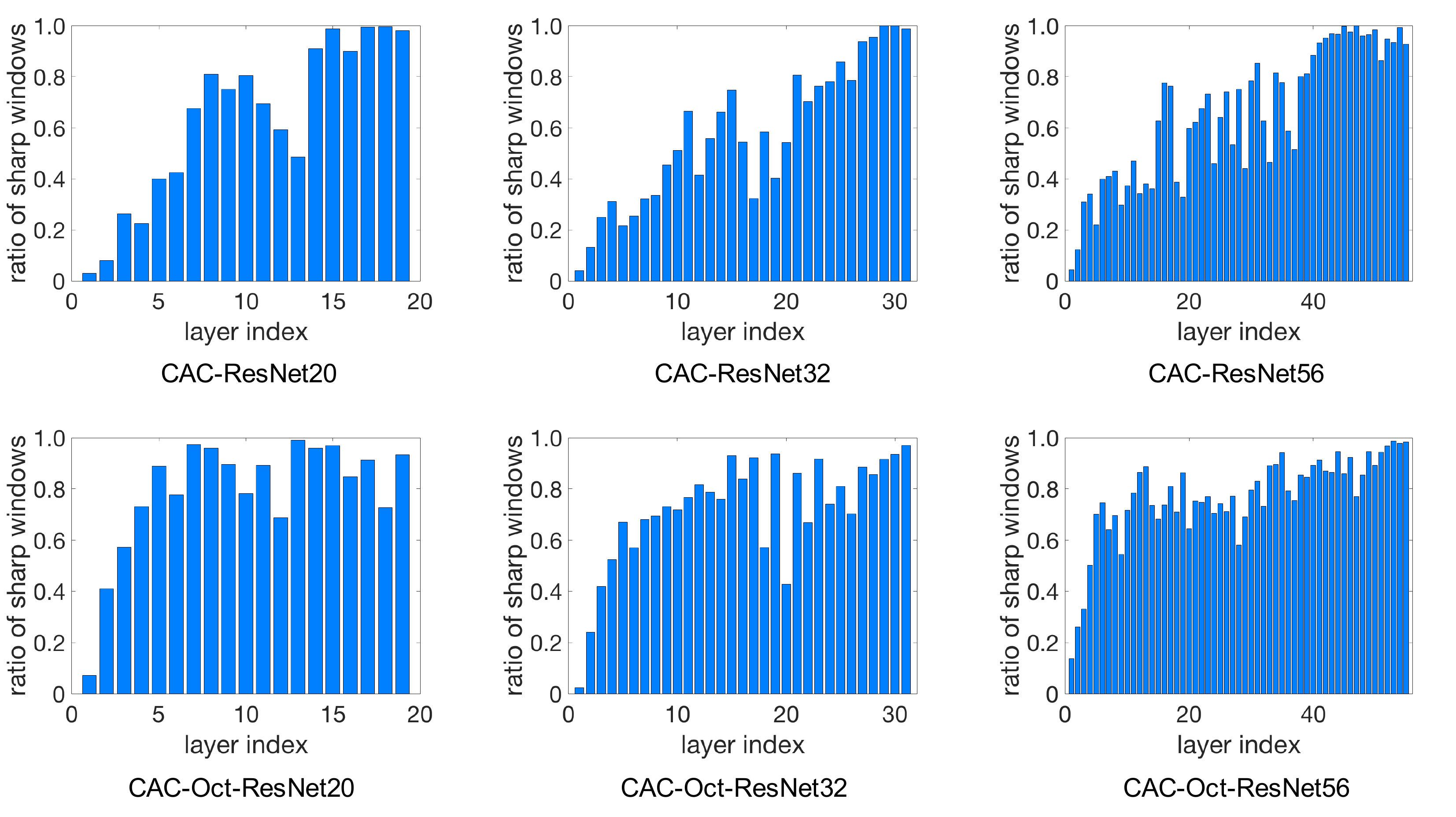}
	\caption{{Visualization of the ratios of sharp windows for different layers in ResNet20, ResNet32, and ResNet56 on CIFAR-10. We compute the ratios by averaging the ratios over 10,000 testing samples.}}
	\label{fig:cifar_arc}
\end{figure*}

In this experiment, we evaluate our CAC method on a small dataset CIFAR-10.
From Table~\ref{tab:cifar}, the proposed CAC method greatly accelerates ResNet and DenseNet models equipped with different convolution types.
Specifically, for both the models with the standard convolution and OctConv, our CAC \cyf{consistently yields significantly better performance and lower computational cost.}
Moreover, we also show the ratios of sharp windows $\rho$ of each layer based on several models in Fig.~\ref{fig:cifar_arc}.
From this figure, deep layers tend to have larger ratios than shallow layers due to their better representation ability and smaller feature map, yielding a smaller risk of containing smooth windows. 
These results show that the proposed CAC removes the redundancy caused by the smooth windows in each layer.

We also consider very compact models, \eg, ShuffleNetV2, which mainly consists of $1 \times 1$ group convolution.
However, since OctConv requires information exchange among groups, it would destroy the computation of group convolution and thus cannot be directly applied to ShuffleNetV2.
Thus, we only compare the performance of the models equipped with the standard convolution and the proposed CAC.
{Even with such a compact model, our CAC further improves the validation accuracy \cyf{and reduces the redundancy in the model.}
These results demonstrate the effectiveness of our method.}

\subsubsection{Comparisons on ImageNet}
We also evaluate our method on a large-scale dataset ImageNet.
Similar to the experiments on CIFAR-10, we apply our CAC to improve both the standard convolution and OctConv.
In this experiment, we consider ResNet, DenseNet and ShuffleNetV2 as the baseline model.
The results are shown in Table~\ref{tab:imagenet}.

\begin{figure*}[t!]
	\centering
	\includegraphics[trim = 1mm 0mm 2mm 0mm,
	clip, width=\textwidth]{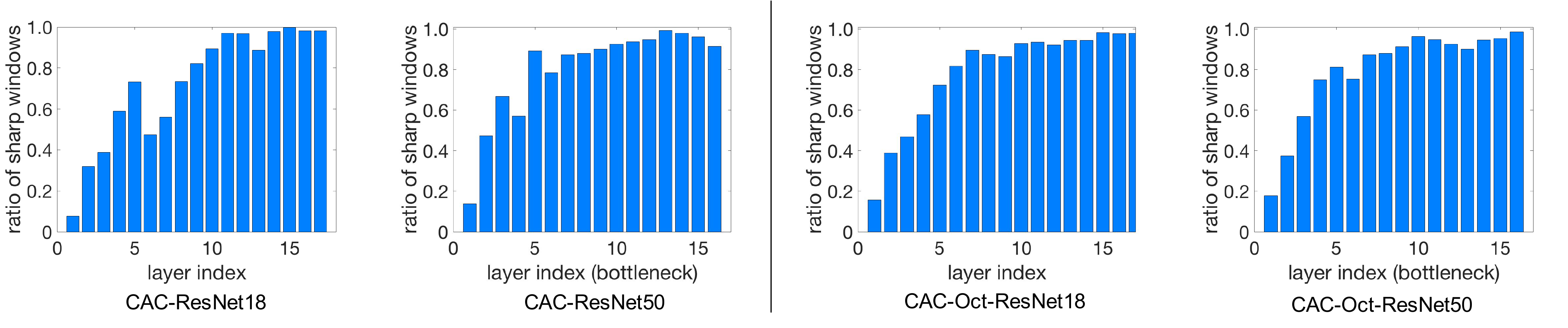}
	\caption{{Visualization of the ratios of sharp windows for different layers in ResNet18 and ResNet50 on ImageNet. The layer index for ResNet50 denotes the index of the bottleneck block. We compute the ratios for different layers by averaging the samples in the validation set.}}
	\label{fig:imagenet_arc}
\end{figure*}

From Table~\ref{tab:imagenet}, our CAC based models significantly outperform the baseline models with different architectures in terms of Top-1 and Top-5 error. More critically, the resultant models often have lower computational cost and thus become more compact. These results demonstrate the superiority of the proposed CAC method over the existing methods.
We also show the ratios $\rho$ of each layer for ImageNet models in Fig.~\ref{fig:imagenet_arc}. 
From this figure, due to the training difficulty on a large-scale dataset, ImageNet models are often hard to compress without performance degradation~\cite{liu2017learning,luo2017thinet,zhuang2018discrimination}, yielding larger ratios of the intermediate layers than the CIFAR-10 models.

\begin{table*}[t!]
  \centering
  \caption{Comparisons of different convolutions in terms of both computational complexity and validation error on ImageNet. ``$\backslash$'' denotes the missing results of the models to which OctConv cannot be applied.}
    \resizebox{\textwidth}{!}
    {
    \begin{tabular}{c||ccc|ccc|ccc|ccc}
    \hline
    \multirow{3}[0]{*}{Conv Type} & \multicolumn{3}{c|}{ResNet18} & \multicolumn{3}{c|}{ResNet50} & \multicolumn{3}{c|}{DenseNet121} & \multicolumn{3}{c}{ShuffleNetV2} \\
     & \multirow{1}[0]{*}{\#MAdds} & \multicolumn{2}{c|}{Error (\%)} & \multirow{1}[0]{*}{\#MAdds} & \multicolumn{2}{c|}{Error (\%)} & \multirow{1}[0]{*}{\#MAdds} & \multicolumn{2}{c|}{Error (\%)} & \multirow{1}[0]{*}{\#MAdds} & \multicolumn{2}{c}{Error (\%)} \\
    \cline{3-4} \cline{6-7} \cline{9-10} \cline{12-13}
    & (G) & Top-1 & Top-5 & (G) & Top-1  & Top-5  & (G) & Top-1 & Top-5 & (G) & Top-1 & Top-5 \\
    \hline
    Conv  & 1.81  & 30.36 & 11.02 & 4.09  & 24.01 & 7.07 & 2.83 & 25.35 & 7.83 & 0.15 & 30.64 & 11.68 \\
    CAC & \textbf{1.41}  & \textbf{30.19} & \textbf{10.87} & \textbf{3.75}  & \textbf{23.79} & \textbf{6.81} & \textbf{2.52} & \textbf{24.49} & \textbf{7.37} & \textbf{0.13} & \textbf{30.13} & \textbf{11.27} \\
    \hline
    OctConv~\cite{chen2019drop} & 1.14  & 29.64 & 10.48 & 2.37  & 23.27 & 6.55 & 1.37 & 25.68 & 7.90 & \multicolumn{3}{c}{\multirow{2}[0]{*}{\diagbox{~~~~~~~~~~~~~}{~~~~~~~~~~~~~}}} \\
    CAC-OctConv & \textbf{0.96}  & \textbf{29.43} & \textbf{10.27} & \textbf{2.19}  & \textbf{23.05} & \textbf{6.37} & \textbf{1.28} & \textbf{24.82} & \textbf{7.61}\\
    \hline
    \end{tabular}%
    }
  \label{tab:imagenet}%
\end{table*}%

\subsection{Experiments on Semantic Segmentation}\label{sec:semantic_segmentation}

We further apply the proposed CAC to semantic segmentation models, \eg, fully convolutional network (FCN)~\cite{long2015fully}.
We compare the performance of the models with and without CAC based on a benchmark dataset PSACAL VOC 2012~\cite{pascal_voc}.

\subsubsection{Compared Methods}
We adopt FCN as the baseline model and apply our CAC to show the effectiveness of CAC. 
In this experiment, we compare our CAC with a strong baseline region convolution (RC).
Moreover, we also consider several semantic segmentation methods as the baselines, including BONN-SVR~\cite{carreira2012object}, $\rm O_2P$~\cite{carreira2012semantic}, SDS~\cite{hariharan2014simultaneous}, and MSRA-CFM~\cite{dai2015convolutional}.

\begin{figure*}[t!]
	\centering
		\includegraphics[width=0.97\textwidth]{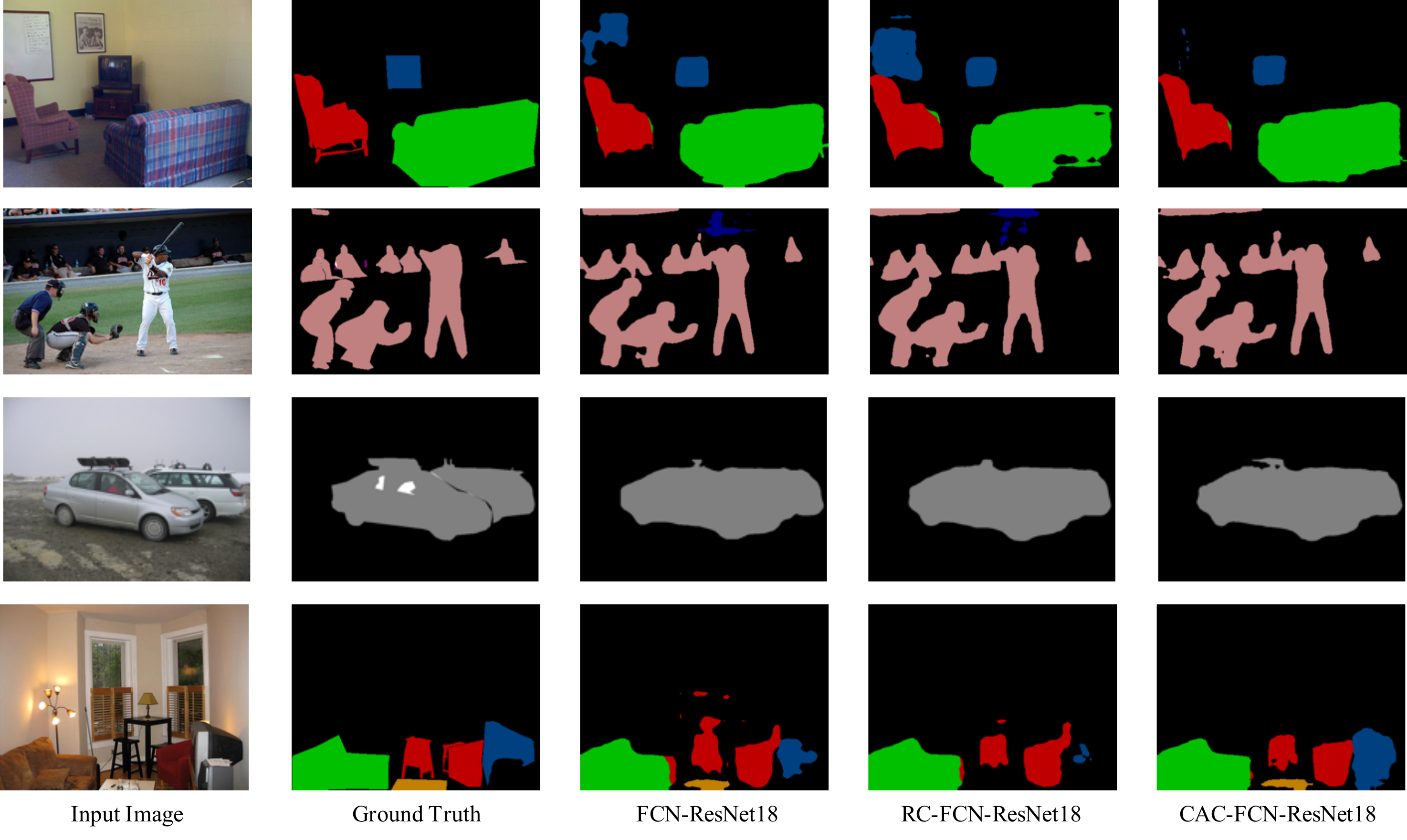}
	\caption{Visual comparison of the segmentation masks produced by different methods. The first and the second columns show the input images and the corresponding ground truth segmentation masks, respectively. The last three columns show the segmentation masks predicted by different models.}
	\label{fig:segmentation_comparison}
\end{figure*}

\subsubsection{Datasets and Implementation Details}

We conduct experiments on the benchmark semantic segmentation dataset PASCAL VOC 2012,
which consists of $1,464$ training images and $1,449$ validation images. 
We measure the performance using the commonly used metric, \ie, the mean intersection over union (mIoU), which computes the percent between the intersection and union of the ground truth segmentation mask and the prediction mask.

In this experiment, we use the ImageNet pretrained model as the backbone model, \eg, ResNet18.
Following the setting in \cite{deeplabv3}, we make some modifications to adapt the model to the semantic segmentation task.
First, we remove the last two subsampling layers in ResNet18 to upscale the size of the output feature map by $4 \times$.
Second, we replace the last four convolution layers with dilated convolutions.
Third, we replace the linear layer with an interpolated upsampling layer.
In the training, we first train the model on MS-COCO dataset and then finetune it on PASCAL VOC 2012.
We make some modifications to adapt it to the semantic segmentation task.
We apply data augmentation by randomly scaling the input images (from 0.5 to 2.0) and randomly left-right flipping in training.
The input images are resized to $480\times480$ in testing.
We finetune 30 epochs using mini-batch SGD with a weight decay of $0.0001$ and a momentum of $0.9$.
The learning rate is started at $0.01$.
In all experiments, we train the CAC based models with $\lambda=0.3$ and $C=0.5$. 

{
\renewcommand\tabcolsep{2.2pt}
\begin{table*}[tbp]
  \centering
  \caption{Comparisons of different models on each class for semantic segmentation. We adopt the FCN-ResNet18 model as the baseline model. ``-'' denotes the results that are not reported.}
  \resizebox{1.01\textwidth}{!}{
    \begin{tabular}{c|cccccccccccccccccccc|cc}
    \hline
    {Method} & aero & bike & bird & boat & bottle & bus & car & cat & chair & cow & table & dog & horse & mbike & person & plant & sheep & sofa & train & tv & mIoU & \#MAdds (G) \\
    \hline
    BONN-SVR~\cite{carreira2012object} & 54.3 & 23.9 & 39.5 & 35.3  & 42.6 & 65.4 & 53.5 & 46.1 & 15.0 & 47.4 & 30.1 & 33.9 & 48.8 & 54.4 & 46.4 & 28.8 & 51.3 & 26.2  & 44.9 & 37.2 & 43.3 \\
    $\rm O_2P$~\cite{carreira2012semantic} & 64.0 & 27.3 & 54.1 & 39.2 & 48.7 & 56.6 & 57.7 & 52.5 & 14.2 & 54.8 & 29.6 & 42.2 & 58.0 & 54.8 & 50.2 & 36.6 & 58.6 & 31.6 & 48.4 & 38.6 & 47.8 & - \\
    SDS~\cite{hariharan2014simultaneous} & 63.3 & 25.7 & 63.0 & 39.8 & 59.2 & 70.9 & 61.4 & 54.9 & 16.8 & 45.0 & 48.2 & 50.5 & 51.0 & 57.7 & 63.3 & 31.8 & 58.7 & 31.2 & 55.7 & 48.5 & 51.6 & - \\
    {MSRA-CFM~\cite{dai2015convolutional}} & 75.7 & 26.7 & 69.5 & 48.8 & 65.6 & 81.0 & 69.2 & 73.3 & 30.0 & 68.7 & 51.5 & 69.1 & 68.1 & 71.7 & 67.5 & 50.4 & 66.5 & 44.4 & 58.9 & 53.5 & 61.8 & - \\
    \hline
    {FCN}~\cite{long2015fully} & 83.5 & 29.7 & 68.7 & 59.8 & 50.2 & 80.3 & 71.9 & \textbf{71.8} & 28.6 & 59.6 & 44.4 & \textbf{61.2} & 59.0 & 66.7 & \textbf{80.3} & 40.8 & \textbf{63.4} & 42.5 & 74.4 & 66.9 & 61.7 & 45.0 \\
    RC-FCN~\cite{li2017not} & 82.8 & 28.9 & 64.4 & 58.6 & 50.3 & 80.5 & 70.5 & 71.6 & 26.7 & 58.3 & 45.5 & 58.4 & 58.1 & 67.8 & 79.6 & 40.9 & 58.6 & 42.7 & 73.9 & 63.6 & 60.6 & 38.2 \\
    CAC-FCN & \textbf{84.1} & \textbf{29.9} & \textbf{69.1} & \textbf{60.5} & \textbf{51.9} & \textbf{82.2} & \textbf{73.1} & 71.5 & \textbf{28.9} & \textbf{59.7} & \textbf{48.0} & 59.5 & \textbf{60.2} & \textbf{69.2} & 80.2 & \textbf{41.8} & 63.1 & \textbf{44.1} & \textbf{76.9} & \textbf{67.0} & \textbf{62.5} & \textbf{37.2}  \\
    \hline
    \end{tabular}%
    }
  \label{tab:segmentation}%
\end{table*}%
}

\subsubsection{Performance Comparison}

{In this section, we compare the proposed CAC method with the standard convolution and region convolution (RC) on semantic segmentation tasks. For convenience, we use FCN, RC-FCN, and CAC-FCN to represent the models with the standard convolution, RC, and CAC. We show the results in Table~\ref{tab:segmentation} and Fig.~\ref{fig:segmentation_comparison}.}

From Table \ref{tab:segmentation}, our CAC-FCN outperforms the baseline models with the standard convolution and RC on most of the categories. For the average performance in terms of mIoU, the CAC-FCN model yields significant performance improvement.
We also provide a visual comparison of different models in Fig.~\ref{fig:segmentation_comparison}.
From this figure, our CAC-FCN produces more accurate segmentation masks than the FCN and RC-FCN baselines. 

\subsection{Experiments on Object Detection}\label{sec:object_detection}

In this section, we apply the proposed CAC method to object detection models. We evaluate the CAC-based models on the benchmark dataset MS COCO~\cite{coco}.

\subsubsection{Compared Methods}
In this experiment, we adopt the widely used model Faster-RCNN network as the baseline model.
We compare our CAC based model with several state-of-the-art object detection models, including Fast-RCNN~\cite{girshick2015fast}, ION~\cite{bell2016inside}, YOLOv2~\cite{redmon2017yolo9000}, SSD300~\cite{liu2016ssd} and SSD512~\cite{liu2016ssd}.

\subsubsection{Datasets and Implementation Details}

We conduct experiments on the MS COCO dataset which contains $117$k training images and $5$k validation images with $80$ classes.
We use the ImageNet pretrained ResNet18-based FPN network for comparing different methods on MS COCO dataset.
We follow the setting in \cite{fasterrcnn}. 
The networks are optimized for 13 epochs using SGD with a weight decay of 0.0001 and a momentum of 0.9.
The learning rate is initialized with $0.02$ and divided by 10 at 8 and 11 epochs.

We evaluate different object detection models using the COCO’s standard metric, namely mAP@0.5 and mAP@0.75. These two metrics represent the mean average precision (mAP) computed at the IoU thresholds of 0.5 and 0.75, respectively.
We also compute mmAP for different models by averaging multiple mAP values with the IoU thresholds ranging from 0.5 to 0.95 with the
step of 0.05.
In all experiments, we train the CAC based models with $\lambda=0.3$ and $C=0.5$.

\begin{table}[t]
  \centering
  \caption{Comparisons of different object detection models on MS COCO dataset. We use the ResNet18 model as the backbone. ``-'' denotes results that are not reported.}
  \resizebox{0.8\textwidth}{!}
  {
    \begin{tabular}{c|cccc}
    \hline
    Model & \multicolumn{1}{l}{mmAP} & \multicolumn{1}{l}{mAP@0.5} & \multicolumn{1}{l}{mAP@0.75} & \#MAdds (G) \\
    \hline
    Fast-RCNN~\cite{girshick2015fast} & 18.9 & 38.6 & - & - \\
    ION~\cite{bell2016inside} & 23.6 & 43.2 & 23.6 & - \\
    YOLOv2~\cite{redmon2017yolo9000} & 21.6 & 44.0 & 27.8 & - \\
    SSD300~\cite{liu2016ssd} & 23.2 & 41.2 & 23.4 & - \\
    SSD512~\cite{liu2016ssd} & 26.8 & 46.5 & 27.8 & - \\
    \hline
    Faster-RCNN~\cite{fasterrcnn} & 28.6 & 49.2 & 29.6 & 23.82 \\
    CAC-Faster-RCNN & \textbf{29.9} & \textbf{49.9} & \textbf{31.1} & \textbf{20.49} \\
    \hline
    \end{tabular}%
    }
  \label{tab:detection}%
\end{table}%

\begin{figure*}[t]
	\centering
		\includegraphics[width=1\textwidth]{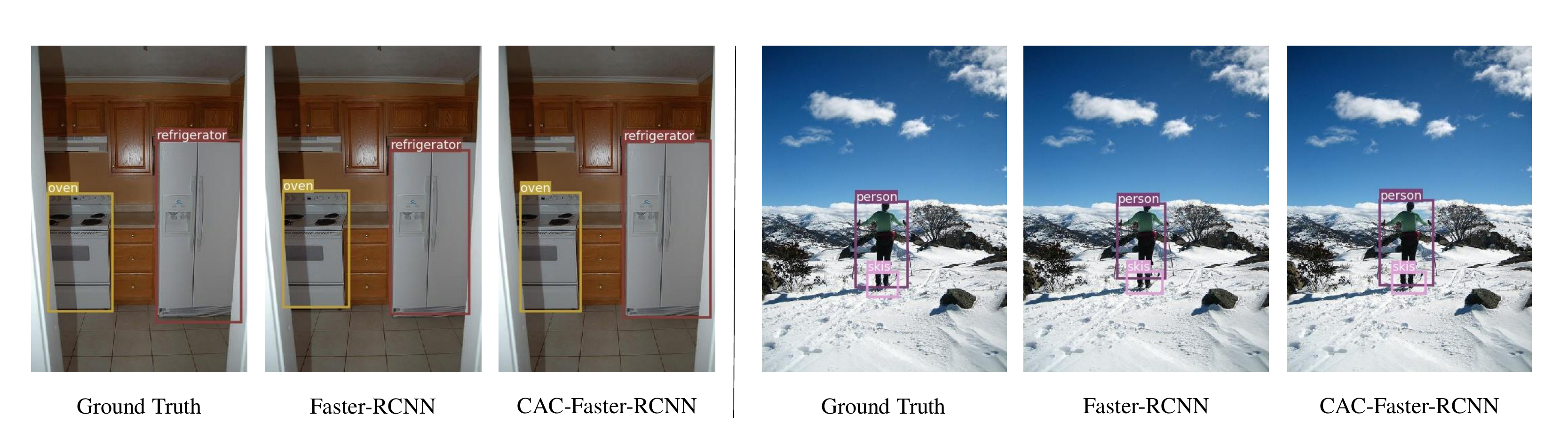}
	\caption{Visual comparison of different object detection methods. The first column shows the input images with the corresponding ground truth bounding boxes. The second and third columns show the bounding boxes predicted by different models.}
	\label{fig:detection_comparison}
\end{figure*}

\subsubsection{Performance Comparison}
In this experiment, we use CAC to replace the standard convolution layers in the Faster-RCNN model. We show the quantitative and visual results in Table~\ref{tab:detection} and Fig.~\ref{fig:detection_comparison}.

From Table \ref{tab:detection}, CAC can obtain very compact models with better performance and lower computational cost than the baseline models equipped with the standard convolution. Specifically, the resultant CAC-Faster-RCNN has a lower computational cost and consistently outperforms the baseline model on all the considered metrics, including mmAP, mAP@0.5, and mAP@0.75. 
We also show the visual comparison of different object detection models in Fig.~\ref{fig:detection_comparison}.
From this figure, our CAC model generates more accurate bounding boxes than the Faster-RCNN baseline.
The results demonstrate the effectiveness of the proposed CAC method on object detection tasks.

\begin{table}[t]
  \centering
  \caption{Comparisons of robustness of deep models with different convolutions on CIFAR-10.}
 \resizebox{1\textwidth}{!}
    {
    \begin{tabular}{c|c|cccc}
    \hline
    \multirow{2}[0]{*}{Model} & \multicolumn{1}{c|}{\multirow{2}[0]{*}{Method}} & \multicolumn{4}{c}{Error on Adversarial Examples (\%)} \\
    \cline{3-6}
          &       & \multicolumn{1}{c}{FGSM~\cite{goodgellow2015explaining}} & \multicolumn{1}{c}{MI-FGSM~\cite{dong2018boosting}} & \multicolumn{1}{c}{PGD-10~\cite{madry2018towards}} & \multicolumn{1}{c}{PGD-100~\cite{madry2018towards}} \\
    \hline
    \multirow{2}[0]{*}{ResNet20} & Standard Conv & 29.82      &  54.89     &   57.99    &  65.41 \\
          & CAC   &   \textbf{28.97}    &    \textbf{53.67}   &   \textbf{57.13}    &  \textbf{64.74} \\
    \hline
    \multirow{2}[0]{*}{ResNet56} & Standard Conv & 26.91      &   51.91    &    55.33   & 64.81 \\
          & CAC   &    \textbf{25.58}   &    \textbf{51.03}   &   \textbf{54.72}    & \textbf{63.75} \\
    \hline
    \end{tabular}%
    }
  \label{tab:robustness}%
\end{table}%

\section{Further Experiments}

\cyf{In this section, we first compare the accuracy of our CAC method with the standard convolution method under adversarial perturbations to investigate the robustness of our method.
Then, we conduct more experiments to investigate the effect of the hyperparameter $\lambda$.}

\subsection{{Comparisons of Robustness}}

\cyf{
We investigate the robustness of the proposed CAC by comparing the accuracy on adversarial samples generated by four different attack methods, including FGSM~\cite{goodgellow2015explaining}, MI-FGSM~\cite{dong2018boosting}, PGD10~\cite{madry2018towards} and PGD100~\cite{madry2018towards}.
We train the standard convolution models (namely ResNet20 and ResNet56) and the CAC based models with adversarial samples on CIFAR-10.
We report the adversarial accuracy that is evaluated on adversarial samples.
The higher adversarial accuracy the model has, the more adversarially robust the model will be.
}

\cyf{
Following the setting in~\cite{madry2018towards}, we train all the models for 200 epochs and use an SGD optimizer with a momentum of 0.9 and a weight decay of 0.0005.
The initialized learning rate is set to 0.1 and divided by 10 at epochs 90, 140 and 160, respectively.
All attacks are with a total perturbation scale of 8/255 (0.03) and a step size of 2/255 (0.01).
We set the number of attack iterations to 10, 10 and 100 for MIFGSM~\cite{dong2018boosting}, PGD10~\cite{madry2018towards} and PGD100~\cite{madry2018towards}, respectively.
}
\cyf{From Table~\ref{tab:robustness}, our CAC based models achieve higher adversarial accuracy than the standard convolution ones under four different attack perturbations.
These results demonstrate that the proposed CAC convolution is more robust than the standard convolution.
The main reason is that our CAC replaces the original large kernel with a $1 \times 1$ kernel to perform convolution on smooth windows. 
In this sense, we are able to effectively reduce the influence incurred by the noises and attacks in these windows.
}

\subsection{Effect of $\lambda$ on CAC}\label{exp:lambda}

We investigate the effect of the hyperparameter $\lambda$ in Eqn.~(\ref{eqn:loss}) on the performance of CAC models.
In this experiment, we use ResNet20 as the baseline model and train the models with different values of $\lambda \in \{0.3, 0.5, 0.7, 1.0 \}$ on CIFAR-10. 
We show the results in Table~\ref{tab:lambda}.

From Table~\ref{tab:lambda}, the reduction of MAdds would increase when we gradually increase $\lambda$.
However, due to the redundancy incurred by the smooth windows, it is possible to simultaneously reduce the computational cost and improve the performance, \eg, when setting $\lambda=0.3$.
If we further increase the value of $\lambda$, the objective in Eqn.~(\ref{eqn:loss}) encourages the model to focus more on the computational cost but compromise the performance.
For example, when $\lambda$ is set to $1.0$, CAC reduces the computational cost by $69.39\%$ but incur significant performance degradation.
Thus, we set $\lambda=0.3$ to train CAC models in the experiments.

\begin{table}[t]
	\centering
	\caption{Comparisons of the CAC-ResNet20 models with different values of $\lambda$ on CIFAR-10. 
	}
    {
	\begin{tabular}{c|c|c|c|c|c}
		\hline
		Method & Standard Conv & \multicolumn{4}{c}{CAC} \\
		\hline
		\multicolumn{1}{c|}{$\lambda$} & - & 0.3   &  0.5   & 0.7   & 1.0\\
		\hline
		Error (\%)   &    8.75   &  \textbf{8.37}   & 9.92 & 10.07 & 11.81 \\
		\#MAdds $\downarrow$ (\%) & 0 &  36.98     &  40.38  &   51.68   & 69.39 \\
		\hline
	\end{tabular}%
	}
	\label{tab:lambda}%
\end{table}%

\section{Conclusion}

In this paper, we have proposed a Content-aware Convolution (CAC) to reduce the computational redundancy incurred by the smooth windows when performing convolution.
To reduce the computational redundancy and improve the performance, CAC replaces the original $k \times k$ kernel with a $1\times 1$ kernel to perform convolutions on the smooth windows.
Moreover, we propose an efficient algorithm to automatically recognize the sharp and smooth windows. 
Given different samples, the resultant CAC models could allocate different computation resources according to their data smoothness, which makes it possible for content-aware computation. 
Extensive results on image classification, semantic segmentation, and object detection tasks show that our CAC based models yield significantly better performance and lower computational cost than the baseline models with the standard convolution.

\section{Acknowledgements}

This work was partially supported by Ministry of Science and Technology Foundation Project (2020AAA0106900), National Natural Science Foundation of China (62072190, 62072186), Key Realm R\&D Program of Guangzhou (202007030007), Fundamental Research Funds for the Central Universities (D2191240), Program for Guangdong Introducing Innovative and Entrepreneurial Teams (2017ZT07X183), Key-Area Research and Development Program of Guangdong Province (2019B010155002), Guangdong Basic and Applied Basic Research Foundation (2019B1515130001), Tencent AI Lab Rhino-Bird Focused Research Program (JR201902), 
Guangzhou Science and Technology Planning Project (201904010197), Opening Project of Guangdong Key Laboratory of Big Data Analysis and Processing.

\bibliography{egbib}

\end{document}